\pdfoutput=1

\documentclass[11pt]{article}

\usepackage[]{EACL2023}

\usepackage{times}
\usepackage{latexsym}

\usepackage[T1]{fontenc}

\usepackage[utf8]{inputenc}

\usepackage{microtype}

\usepackage{inconsolata}

\usepackage{amsmath}
\usepackage{mathtools}
\usepackage{cleveref}
\usepackage{xspace}
\usepackage{fontawesome5}

\newcommand{\blockcomment}[1]{}
\newcommand{\ie}{\emph{i.e.}\xspace}
\newcommand{\eg}{\emph{e.g.}\xspace}
\DeclareMathOperator*{\argmax}{arg\,max}

\title{Interpreting Predictive Probabilities:\\ Model Confidence or Human Label Variation?}

\author{Joris Baan\textsuperscript{\faBicycle}, Raquel Fern{\'a}ndez\textsuperscript{\faBicycle }, Barbara Plank\textsuperscript{\faMountain\faCompass\faCar}, Wilker Aziz\textsuperscript{\faBicycle} \\
        {\footnotesize \faBicycle}University of Amsterdam, 
        {\scriptsize \faCompass}IT University of Copenhagen,
        {\scriptsize \faMountain}MCML Munich,
        {\scriptsize \faCar}LMU Munich
        \\
        \texttt{\{j.s.baan,raquel.fernandez,w.aziz,\}@uva.nl},  \texttt{b.plank@lmu.de}}

\begin{document}
\maketitle

\begin{abstract}
With the rise of increasingly powerful and user-facing NLP systems, there is growing interest in assessing whether they have a good \textit{representation of uncertainty} by evaluating the quality of their predictive distribution over outcomes. We identify two main perspectives that drive starkly different evaluation protocols. 
The first treats predictive probability as an indication of model confidence; 
the second as an indication of human label variation.
We discuss their merits and limitations, and take the position that both are crucial for trustworthy and fair NLP systems, but that exploiting a single predictive distribution is limiting. We recommend tools and highlight exciting directions towards models with disentangled representations of uncertainty about predictions and uncertainty about human labels.
\end{abstract}

\section{Introduction}
\label{sec:introduction}
In common language, uncertainty refers to ``a state of not being definitely known or perfectly clear; a state of doubt''.\footnote{Oxford English Dictionary, accessed October 13th 2023.} 
In statistics and machine learning, uncertainty is taken as a state to be represented \cite{lindley2013understanding, halpern2017reasoning}---the state of the world as a function of inherently stochastic experiments or the state of knowledge of an agent observing or interacting with the world---and its mathematical representation requires prescribing a probability measure \cite{kolmogorov1960foundations}. 

In modern NLP, neural networks are the de-facto standard to predict complex probability measures from available context \citep{neural-goldberg-hirst-2017}: given an input (or prompt), a neural network prescribes a representation of uncertainty over the space of responses (\eg, strings or classes), typically, by mapping the input to the parameter of a probability mass function (\eg, in text classification, inputs are mapped to the probability masses of each outcome in the label space). 

Recently, transformer-based large language models (LLMs) are becoming increasingly powerful and display remarkable abilities on complex classification tasks, leading to an increased deployment in user-facing applications. This motivates the need for models that can signal when they are likely to be wrong (\textbf{P1}; an aspect of trustworthiness), and models that can capture different linguistic and human interpretations (\textbf{P2}; an aspect of language including fairness).

In this position paper, we identify that the exact same representation of uncertainty---the predictive distribution over outcomes---is sometimes interpreted as an indication of confidence in model predictions \cite[\textbf{P1;}][]{desai-durrett-2020-calibration,dan-roth-2021-effects-transformer,jiang2021can} and other times as an indication of variation in human perspectives \cite[\textbf{P2;}][]{plank-2022-problem}.

We hope to provide clarity and accelerate progress by:

(i) Identifying these two perspectives on the predictive distribution and examining how each evaluates the quality of predictive distribution in \Cref{sec:interpretations}. 

(ii) Discussing their merits and limitations, and relating them to popular notions of \textit{aleatoric} and \textit{epistemic} uncertainty in \Cref{sec:merits}.

(iii) Taking the position that both perspectives contribute to trustworthy and fair NLP systems, but that exploiting a single predictive distribution is limiting---\eg, does a uniform predictive distribution represent uncertainty about human perspectives, or rather about the correctness of that prediction itself?---and highlighting exciting directions towards models that can predict distributions over human or linguistic interpretations, and simultaneously abstain from answering when lacking such knowledge or skills in \Cref{sec:directions}.

\section{Two Perspectives on Uncertainty}
\label{sec:interpretations}

Consider a user-facing question answering (QA) system. Ideally, this model is able to abstain on questions that it is likely to get wrong \citep[a.k.a. selective answering or prediction; ][]{kamath-etal-2020-selective, yoshikawa2023selective}, for which its probabilities should reflect confidence in predictions (\ie, predictive probabilities help us determine whether the model is right or wrong). 
Now consider that various NLP tasks, including QA, are being acknowledged as supporting human label variation \citep{plank-2022-problem}, and that some questions can be underspecified, ambiguous or subjective (there are many such datasets, for QA see for example \citet{min-etal-2020-ambigqa} and \citet{amouyal2023qampari}, and for other tasks see Section \ref{subsec:merits_data}). Different annotators might therefore provide a different reference answer. From this perspective,  probabilities should reflect the relative frequency of each answer assigned to that particular question by the pool of annotators (\ie, predictive probabilities help us determine what answers represent the views of a certain population).
These two perspectives on the role of predictive probabilities  in fact aim at different sources of uncertainty: uncertainty about model error (\eg, due to imperfect design and estimation) and uncertainty about human labels (\eg, due to label variation in a population). So, if a model predicts a uniform distribution, does this mean that all answers are plausible or that this prediction should not be trusted?

\subsection{Background}

Most text classifiers chain two  building blocks: i) a parametric model which, given input text $x$, prescribes the probability mass function (pmf) $f(y; x)$ of the conditional random variable $Y|X=x$ taking on values in a set $\{1, \ldots, K\}$ of $K$ class labels; and ii) a decision rule $\delta_f(x)$ to map from $f(\cdot; x)$ to a single label. 
For most modern models, the map $x \mapsto f(\cdot; x)$ is realised by a neural network  and the most common decision rule $\delta_f(x) = \argmax_{k \in [K]}~f(k; x)$  returns the mode of the pmf. Next, we identify two main perspectives on predictive probability $f(y;x)$, with starkly different evaluation frameworks.\footnote{We use capital letters for random variables (\eg, $X$, $Y$) and lowercase letters for outcomes (\eg, $x$, $y$). As standard, $X=x$ denotes random variable (rv) assignment. For logical predicates we use the Iverson bracket $[A=B]$ to denote a new rv whose outcome is $1$, when $A$ and $B$ are assigned the same outcome, and $0$ otherwise. A deterministic function of an rv defines a new rv; for example, the rv $\delta_f(X) = \argmax_{k \in [K]}~f(k; X)$ captures the mode of the conditional distribution as a function of the random input $X$. We use $\Pr$ to denote an implicit probability measure capturing the data generation process; we do not possess an explicit representation for this measure, but we can estimate its assessment via Monte Carlo--that is, the relative frequency of the relevant events in a dataset of labelled inputs.} %

\subsection{P1: Uncertainty about Model Error}
\label{subsec:calibration}
The first and arguably more common perspective interprets predictive probabilities as predictive of \textit{classification performance} and is often explained as evaluating the extent to which ``a model knows when it does not know'' \cite[\eg, in NLP:][]{desai-durrett-2020-calibration, dan-roth-2021-effects-transformer,jiang2021can}. An increasingly popular evaluation framework taking this perspective is calibration.

The core desideratum behind \textit{confidence calibration} \cite{naeini2015obtaining, guo2017calibration} is that, \textbf{in expectation over inputs}, a classifier's predictive mode probability $\pi_f(X) = \max_{k \in [K]}~f(k; X)$ matches the relative frequency of predictions $\delta_f(X) = \argmax_{k \in [K]}~f(k; X)$ being judged as correct $[Y = \delta_f(X)]=1$. So, 
 $\forall q \in [0,1],$  
\begin{multline}
    \label{eq:calibration}
    \Pr\left( [Y = \delta_f(X)]=1 \mid \pi_f(X) = q\right) \stackrel{?}{=}q ~.  
\end{multline}
For example, if 100 predictions are made with probability $0.9$, then 90 should be judged as correct.\footnote{Other notions assess calibration for fixed classes \citep[\textit{class-wise};][]{nixon2019measuring}
or probability vectors \citep[\textit{multi-class};][]{vaicenavicius2019evaluating,kull2019beyond}.} 
In practice \Cref{eq:calibration} is hard to MC estimate (for it requires observing multiple predictions with identical probability), so the probability space is partitioned into $M$ bins. For each bin $B_m$, the calibration error is the difference between accuracy and average probability of the predictions in it. The expected calibration error (ECE) is the weighted average over bins:
\begin{align}
    \label{eq:ece}
    \mathrm{ECE} &= \sum_{m=1}^M\frac{|B_m|}{N}(\mathrm{acc}(B_m)-\mathrm{conf}(B_m)) ~.
\end{align}

\subsection{P2: Uncertainty about Human Labels} 
\label{subsec:fitness}
Crucially, the above interpretation is different from evaluating, \textbf{for each individual input $x$}, whether the predictive probability $f(k;x)$ matches the relative frequency with which (a population of) humans would pick that same label $k$: $\forall k \in [K],$
\begin{equation}
    \label{eq:fitness}
    \Pr(Y=k|X=x) \stackrel{?}{=} f(k; x)  ~.
\end{equation}

Although there is no standard evaluation protocol yet \citep{lovchinsky2020discrepancy,basile-etal-2021-need,plank-2022-problem}, researchers use datasets with multiple annotations per input to estimate a \textit{human distribution}, 
and compare that to the predictive distribution through statistical divergence (\eg, Kullback-Leibner or Jensen-Shannon Divergence; Total Variation Distance), or summary statistics like entropy  \citep{pavlick-kwiatkowski-2019-inherent,nie-etal-2020-learn,baan-etal-2022-stop,glockner2023ambifc}.

\subsection{Ambiguity in Explaining Calibration}
The language that is often used to explain calibration allows (quite ironically) for both perspectives \textbf{P1} and \textbf{P2}. 
\citet{desai-durrett-2020-calibration}: ``\textit{If a model assigns 70\% probability to an event, the event should occur 70\% of the time if the model is calibrated}''. 
The word ``event'' can refer to observing a class given an input (\textbf{P2}) or a model prediction matching the observed class (\textbf{P1}). 

\citet{jiang-etal-2021-know}: ``\textit{the property of a probabilistic model’s predictive probabilities actually being well correlated with the probabilities of correctness}''. The word ``correctness'' can refer to the probability of observing that class in the data (\textbf{P2}) or to the probability of a predicted class matching the data (\textbf{P1}).

\citet{gupta2021calibration}: ``\textit{a classifier is said to be calibrated if the probability values it associates with the class labels match the true probabilities of correct class assignments}'' and ``\textit{It would be desirable if the numbers $z_k$ output by a network represented true probabilities}''. Human annotators could assign the class (\textbf{P2}), or a model could (\textbf{P1}). The phrase ``true probabilities'' could refer to observed class (\textbf{P2}) or model error (\textbf{P1}) frequencies.

The examples above illustrate well that one may regard predictive probabilities one way or another, each interpretation tracking a different type of event (\ie,  correctness, assessed marginally for a collection of inputs, or label frequency, assessed conditionally against a population of annotators). 
Crucially, however, most models are trained to approximately recover the maximum likelihood solution---a single realisation of the map $x \mapsto f(\cdot;x)$, with no room for quantification of uncertainty about its correctness. Therefore, without special incentives (\eg, regularisation, change of loss or supervision; some of which we discuss in Section \ref{subsec:trustworthy}), our predictive distributions are not meant to inherently support \textbf{P1}, and they \emph{may} support \textbf{P2}, as we discuss in the next section.

\section{Merits and Limitations}
\label{sec:merits}
The predictive distribution for an input $x$ is sometimes taken as a representation of uncertainty about \textbf{a model's future classification performance} (``knowing when it knows''); other times as a representation of uncertainty about \textbf{label frequency in a population of human annotators} (human label variation). 
We now discuss merits and limitations for each perspective. 

\subsection{P1: Uncertainty about Model Error}
\label{subsec:merits_calibration}
From a statistical perspective, most NLP systems are trained on single annotations using regularised maximum likelihood estimation (MLE), without mechanism or  incentive to represent uncertainty about their own correctness (MLE recovers a single realisation of the map $x \mapsto f(\cdot; x)$). This is unlike, for instance, Bayesian estimation (where the map $x \mapsto f(\cdot; x)$ is given random treatment; more in \Cref{sec:directions}).
\looseness-1

In addition, regardless of whether \textit{models} represent uncertainty about their own correctness, 
calibration \textit{metrics}, and ECE in particular, are known to have limitations, e.g., problems with binning \cite{nixon2019measuring,vaicenavicius2019evaluating,gupta2021calibration}, evaluating only the mode probability rather than the entire distribution \cite{kumar2019verified, vaicenavicius2019evaluating,widmann2019calibration,kull2019beyond}, and being minimised by global label frequencies \cite{nixon2019measuring}. Moreover, \citet{baan-etal-2022-stop} recently demonstrate that ECE disregards plausible instance-level label variation and pose that such calibration metrics are ill-suited for tasks with human label variation.

Finally, the sense of trustworthiness from verifying that \Cref{eq:calibration} holds (for a given confidence level $q$) in a given dataset, might not transfer to any one future prediction in isolation. 
Though some studies examine the effect of communicating predictive probability to human decision makers \citep{zhang-etal-2020-effect,wang-yin-2021-explanations,vodrahalli2022uncalibrated,vasconcelos2023generation, dhuliawala-etal-2023-diachronic}, to the best of our knowledge, none verified the user-impact of models with various calibration scores, raising the question: can calibration metrics like ECE discriminate systems perceived as more trustworthy?

\subsection{P2: Uncertainty about Human Labels}
\label{subsec:merits_data}
The idea that gold labels are too simplistic has been around for some time \cite{poesio-artstein-2005-reliability,aroyo2015truth} and is gaining traction with increasing evidence that annotators can plausibly pick different class labels for an input \citep{plank-2022-problem}. Examples include subjective tasks such as hate speech detection \cite{kennedy2022introducing} and textual emotion recognition \cite{demszky-etal-2020-goemotions}; and ambiguous or difficult tasks like object naming \cite{silberer-etal-2020-humans}, textual entailment \cite{pavlick-kwiatkowski-2019-inherent, nie-etal-2020-learn}, part-of-speech tagging \cite{manning2011part,plank-etal-2014-linguistically} and discourse relation classification \cite{scholman-etal-2022-discogem}. However, the connection to uncertainty is relatively new \citep{pavlick-kwiatkowski-2019-inherent,nie-etal-2020-learn,baan-etal-2022-stop}. 

From a statistical perspective, text classifiers predict a distribution for $Y|X=x$, and are \textit{precisely} mechanisms to represent uncertainty about a given input's label. However, given that they are parametric models trained with regularised MLE, they can at best learn to predict \textit{observed} label variability (which is often not present in NLP datasets since most record only single annotations), or label variability as a \textit{byproduct} of parametric bottlenecks, regularisation and other inductive biases that reserve (conditional) probability for unseen labels.

Evaluating whether probability mass is indeed allocated coherently with plausible variability is limited by: 1) datasets lacking multiple high quality annotations per input, 2) unclarity about how many annotations are sufficient to reliably estimate the human distribution \cite{zhang-etal-2021-learning-different}, 3) how to separate plausible variation from noise---for example due to spammers \citep{raykar2011ranking,beigman-klebanov-beigman-2014-difficult,aroyo-etal-2019-crowdsourcing}, and 4) the assumption of one unique human distribution being a simplification: subpopulations can cause the marginal distribution not to be representative of its individual components \cite{baan-etal-2022-stop,jiang2023understanding}.

\subsection{Sources of Uncertainty}
These two perspectives on the predictive distribution in NLP can be put in a broader context of statistics and machine learning by considering that there can be many sources that lead to uncertainty \cite{der2009aleatory,hullermeier2021aleatoric,gruber2023sources,jiang2023understanding, baan2023uncertainty}. For example, underspecified input, ambiguity, noise or lack of training data can all be considered sources that may lead to uncertainty.

Such sources are often categorised as \textit{aleatoric} (irreducible; inherent to data) or \textit{epistemic} (reducible, inherent to modelling). In that sense, \textbf{P1} regards the predictive distribution as epistemic uncertainty, whereas \textbf{P2} as aleatoric uncertainty. Armed with this knowledge, one can pick the right modeling tools for each, and tap into this broader literature. In the next section, we make several recommendations.

\section{Best of Both Worlds}
\label{sec:directions}
We argue that the desiderata behind both perspectives are equally important for trustworthy and fair NLP systems, but that expecting the predictive distribution to represent both is limiting. 
Rather than calibrating the predictive distribution to better indicate model error, we outline alternative directions to capture uncertainty about predictions (towards more trustworthy NLP) \textit{and} uncertainty about human perspectives (towards fairer NLP)---where the latter can, and in our view \textit{should} be represented by the predictive distribution.\looseness-1

\subsection{Towards More Trustworthy NLP Systems}\label{subsec:trustworthy}
Inspired by machine translation quality estimation \cite[\eg][]{blatz-etal-2004-confidence,specia-etal-2009-estimating,fomicheva-etal-2020-unsupervised} and the observation that models fail in predictable ways, one could train a (separate) module to predict errors. Ideally, this module is  uncertainty-aware \citep{glushkova-etal-2021-uncertainty-aware}, and  predicts fine-grained errors \citep{dou-etal-2022-gpt}. Predictive probabilities (or summaries like entropy) are features that can be combined with, for example, model explainability features \cite{li-etal-2022-calibration,ye-durrett-2022-explanations,park-caragea-2022-calibration} or input properties \cite{dong-etal-2018-confidence, kamath-etal-2020-selective}.

Alternatively, the event space can be expanded beyond only the target variable to include parameters too, thus allowing for uncertainty about them. Since this leads to intractability, some (approximate) Bayesian solutions in NLP include Langevin dynamics \citep{gan-etal-2017-scalable,shareghi-etal-2019-bayesian}, Monte Carlo dropout \citep{shelmanov-etal-2021-certain, vazhentsev-etal-2022-uncertainty}, ensembling \citep{ulmer-etal-2022-exploring}, variational inference \cite{ponti-etal-2021-parameter}, and stochastic attention \citep{pei2022transformer}. Other directions rely on the distance of a new input to the training data, like conformal prediction \citep{pmlr-v128-maltoudoglou20a,pmlr-v152-giovannotti21a,zerva2023conformalizing} or feature space density \cite{van2020uncertainty,vazhentsev-etal-2022-uncertainty,mukhoti2023deep}.

Evaluating model error uncertainty is challenging, in part  because ground truth is difficult to find. Proxy tasks like selective answering  \citep{dong-etal-2018-confidence,kamath-etal-2020-selective, yoshikawa2023selective} are useful due to their flexibility in defining quality (other than accuracy), and error indicators (other than predictive probability), and we encourage more principled evaluation methods.

\citet{rottger-etal-2022-two} propose two annotation paradigms: encouraging the \textit{description} of multiple beliefs or \textit{prescription} of one consistent belief. Prescriptive datasets, by definition, have no data uncertainty, and although that does not change merits of the model-error perspective, one could now safely supervise models to be more coherent with this interpretation (the goal of calibration), \eg by minimising ECE directly, or through other regularisation objectives \citep{kong-etal-2020-calibrated}.

\subsection{Towards Fairer NLP Systems}
To represent uncertainty about plausible human interpretations, data is crucial. For example: how are annotators recruited, what are their backgrounds, how diverse is the population, what guidelines do they follow, what is their incentive, how focused are they, what is their prior experience or expertise, how many annotations per input are collected? 

In NLP, these factors are commonly not controlled for. However, recently, researchers use annotator information to model sub-populations \cite{al-kuwatly-etal-2020-identifying,akhtar2020modeling} or even individual annotators \cite{geva-etal-2019-modeling,davani-etal-2022-dealing,gordon2022jury}. Without access to such information, others collect and train on multiple annotations per instance \citep{peterson2019human,uma2020case,fornaciari-etal-2021-beyond,uma2021learning,zhang-etal-2021-learning-different,Meissner-etal-2021-embracing}, or individual annotator confidence scores \citep{chen-etal-2020-uncertain,collins2022eliciting}. 

Besides data, an appealing but non-trivial alternative (for some tasks, like textual entailment) is to encourage models to generalise to the linguistic phenomena that give rise to label variation, despite supervising with single annotations \citet{pavlick-kwiatkowski-2019-inherent}. Yet another direction is to isolate and understand specific sources of label variation, for example, linguistic ambiguity, and design targeted methods to model them \citep{beck-etal-2014-joint,jiang2022investigating,liu2023we}.

Not all variability is desirable. However, detecting or even defining annotation errors when variation is plausible is difficult. Annotation error detection methods exist, however currently focus on gold labels \citep{wei2022learning,klie2022annotation,weber-plank-2023-activeaed}. We encourage studying noise in label variation settings \citep{paun-etal-2018-comparing,gordon-etal-2021-disagreement}.

\section{Conclusion}
In this position paper, we identified two important perspectives on the predictive distribution in NLP.
We believe that the desiderata behind both are crucial for fair and trustworthy NLP systems, but that exploiting the same predictive distribution is limiting. We recommend exiting tools and directions to represent uncertainty about predictions (model confidence) and about label variation (human perspectives). We hope to facilitate a better understanding of uncertainty in NLP, and encourage future work to acknowledge, represent and evaluate multiple sources of uncertainty with principled design decisions.\looseness=-1

\section*{Limitations}
Evaluation along a specific axis can be useful regardless of whether a model has been explicitly designed to meet this goal. One could argue this is true for both calibration as well as human label variation. It is certainly also true in other sub-fields, like interpretability. For example, probing hidden representations or specific linguistic information, without having explicitly trained models to store them.
Furthermore, although we focus on classification systems in the language domain, the topics we highlight and discuss are equally important in other domains, such as computer vision (\eg, affective computing), or language generation (\eg, story telling).

\section*{Acknowledgements} 
JB is supported by the ELLIS Amsterdam Unit. RF and BP are supported by the European Research Council (ERC), grant agreements No.\ 819455 (DREAM) and No.\ 101043235 (DIALECT), respectively. WA is supported by the EU's Horizon Europe research and innovation programme (grant agreement No.\ 101070631, UTTER).

\bibliography{anthology,anthology_p2, custom}

\begin{thebibliography}{92}
\expandafter\ifx\csname natexlab\endcsname\relax\def\natexlab#1{#1}\fi

\bibitem[{Akhtar et~al.(2020)Akhtar, Basile, and Patti}]{akhtar2020modeling}
Sohail Akhtar, Valerio Basile, and Viviana Patti. 2020.
\newblock Modeling annotator perspective and polarized opinions to improve hate
  speech detection.
\newblock In \emph{Proceedings of the AAAI Conference on Human Computation and
  Crowdsourcing}, volume~8, pages 151--154.

\bibitem[{Al~Kuwatly et~al.(2020)Al~Kuwatly, Wich, and
  Groh}]{al-kuwatly-etal-2020-identifying}
Hala Al~Kuwatly, Maximilian Wich, and Georg Groh. 2020.
\newblock \href {https://doi.org/10.18653/v1/2020.alw-1.21} {Identifying and
  measuring annotator bias based on annotators{'} demographic characteristics}.
\newblock In \emph{Proceedings of the Fourth Workshop on Online Abuse and
  Harms}, pages 184--190, Online. Association for Computational Linguistics.

\bibitem[{Amouyal et~al.(2023)Amouyal, Wolfson, Rubin, Yoran, Herzig, and
  Berant}]{amouyal2023qampari}
Samuel~Joseph Amouyal, Tomer Wolfson, Ohad Rubin, Ori Yoran, Jonathan Herzig,
  and Jonathan Berant. 2023.
\newblock \href {http://arxiv.org/abs/2205.12665} {Qampari: An open-domain
  question answering benchmark for questions with many answers from multiple
  paragraphs}.

\bibitem[{Aroyo et~al.(2019)Aroyo, Dixon, Thain, Redfield, and
  Rosen}]{aroyo-etal-2019-crowdsourcing}
Lora Aroyo, Lucas Dixon, Nithum Thain, Olivia Redfield, and Rachel Rosen. 2019.
\newblock \href {https://doi.org/10.1145/3308560.3317083} {Crowdsourcing
  subjective tasks: The case study of understanding toxicity in online
  discussions}.
\newblock In \emph{Companion Proceedings of The 2019 World Wide Web
  Conference}, WWW '19, page 1100–1105, New York, NY, USA. Association for
  Computing Machinery.

\bibitem[{Aroyo and Welty(2015)}]{aroyo2015truth}
Lora Aroyo and Chris Welty. 2015.
\newblock Truth is a lie: Crowd truth and the seven myths of human annotation.
\newblock \emph{AI Magazine}, 36(1):15--24.

\bibitem[{Baan et~al.(2022)Baan, Aziz, Plank, and
  Fernandez}]{baan-etal-2022-stop}
Joris Baan, Wilker Aziz, Barbara Plank, and Raquel Fernandez. 2022.
\newblock \href {https://doi.org/10.18653/v1/2022.emnlp-main.124} {Stop
  measuring calibration when humans disagree}.
\newblock In \emph{Proceedings of the 2022 Conference on Empirical Methods in
  Natural Language Processing}, pages 1892--1915, Abu Dhabi, United Arab
  Emirates. Association for Computational Linguistics.

\bibitem[{Baan et~al.(2023)Baan, Daheim, Ilia, Ulmer, Li, Fern{\'a}ndez, Plank,
  Sennrich, Zerva, and Aziz}]{baan2023uncertainty}
Joris Baan, Nico Daheim, Evgenia Ilia, Dennis Ulmer, Haau-Sing Li, Raquel
  Fern{\'a}ndez, Barbara Plank, Rico Sennrich, Chrysoula Zerva, and Wilker
  Aziz. 2023.
\newblock Uncertainty in natural language generation: From theory to
  applications.
\newblock \emph{arXiv preprint arXiv:2307.15703}.

\bibitem[{Basile et~al.(2021)Basile, Fell, Fornaciari, Hovy, Paun, Plank,
  Poesio, and Uma}]{basile-etal-2021-need}
Valerio Basile, Michael Fell, Tommaso Fornaciari, Dirk Hovy, Silviu Paun,
  Barbara Plank, Massimo Poesio, and Alexandra Uma. 2021.
\newblock \href {https://doi.org/10.18653/v1/2021.bppf-1.3} {We need to
  consider disagreement in evaluation}.
\newblock In \emph{Proceedings of the 1st Workshop on Benchmarking: Past,
  Present and Future}, pages 15--21, Online. Association for Computational
  Linguistics.

\bibitem[{Beck et~al.(2014)Beck, Cohn, and Specia}]{beck-etal-2014-joint}
Daniel Beck, Trevor Cohn, and Lucia Specia. 2014.
\newblock \href {https://doi.org/10.3115/v1/D14-1190} {Joint emotion analysis
  via multi-task {G}aussian processes}.
\newblock In \emph{Proceedings of the 2014 Conference on Empirical Methods in
  Natural Language Processing ({EMNLP})}, pages 1798--1803, Doha, Qatar.
  Association for Computational Linguistics.

\bibitem[{Beigman~Klebanov and
  Beigman(2014)}]{beigman-klebanov-beigman-2014-difficult}
Beata Beigman~Klebanov and Eyal Beigman. 2014.
\newblock \href {https://doi.org/10.3115/v1/P14-2064} {Difficult cases: From
  data to learning, and back}.
\newblock In \emph{Proceedings of the 52nd Annual Meeting of the Association
  for Computational Linguistics (Volume 2: Short Papers)}, pages 390--396,
  Baltimore, Maryland. Association for Computational Linguistics.

\bibitem[{Blatz et~al.(2004)Blatz, Fitzgerald, Foster, Gandrabur, Goutte,
  Kulesza, Sanchis, and Ueffing}]{blatz-etal-2004-confidence}
John Blatz, Erin Fitzgerald, George Foster, Simona Gandrabur, Cyril Goutte,
  Alex Kulesza, Alberto Sanchis, and Nicola Ueffing. 2004.
\newblock \href {https://aclanthology.org/C04-1046} {Confidence estimation for
  machine translation}.
\newblock In \emph{{COLING} 2004: Proceedings of the 20th International
  Conference on Computational Linguistics}, pages 315--321, Geneva,
  Switzerland. COLING.

\bibitem[{Chen et~al.(2020)Chen, Jiang, Poliak, Sakaguchi, and
  Van~Durme}]{chen-etal-2020-uncertain}
Tongfei Chen, Zhengping Jiang, Adam Poliak, Keisuke Sakaguchi, and Benjamin
  Van~Durme. 2020.
\newblock \href {https://doi.org/10.18653/v1/2020.acl-main.774} {Uncertain
  natural language inference}.
\newblock In \emph{Proceedings of the 58th Annual Meeting of the Association
  for Computational Linguistics}, pages 8772--8779, Online. Association for
  Computational Linguistics.

\bibitem[{Collins et~al.(2022)Collins, Bhatt, and
  Weller}]{collins2022eliciting}
Katherine~M Collins, Umang Bhatt, and Adrian Weller. 2022.
\newblock Eliciting and learning with soft labels from every annotator.
\newblock In \emph{Proceedings of the AAAI Conference on Human Computation and
  Crowdsourcing}, volume~10, pages 40--52.

\bibitem[{Dan and Roth(2021)}]{dan-roth-2021-effects-transformer}
Soham Dan and Dan Roth. 2021.
\newblock \href {https://doi.org/10.18653/v1/2021.findings-emnlp.180} {On the
  effects of transformer size on in- and out-of-domain calibration}.
\newblock In \emph{Findings of the Association for Computational Linguistics:
  EMNLP 2021}, pages 2096--2101, Punta Cana, Dominican Republic. Association
  for Computational Linguistics.

\bibitem[{Demszky et~al.(2020)Demszky, Movshovitz-Attias, Ko, Cowen, Nemade,
  and Ravi}]{demszky-etal-2020-goemotions}
Dorottya Demszky, Dana Movshovitz-Attias, Jeongwoo Ko, Alan Cowen, Gaurav
  Nemade, and Sujith Ravi. 2020.
\newblock \href {https://doi.org/10.18653/v1/2020.acl-main.372}
  {{G}o{E}motions: A dataset of fine-grained emotions}.
\newblock In \emph{Proceedings of the 58th Annual Meeting of the Association
  for Computational Linguistics}, pages 4040--4054, Online. Association for
  Computational Linguistics.

\bibitem[{Der~Kiureghian and Ditlevsen(2009)}]{der2009aleatory}
Armen Der~Kiureghian and Ove Ditlevsen. 2009.
\newblock Aleatory or epistemic? does it matter?
\newblock \emph{Structural safety}, 31(2):105--112.

\bibitem[{Desai and Durrett(2020)}]{desai-durrett-2020-calibration}
Shrey Desai and Greg Durrett. 2020.
\newblock \href {https://doi.org/10.18653/v1/2020.emnlp-main.21} {Calibration
  of pre-trained transformers}.
\newblock In \emph{Proceedings of the 2020 Conference on Empirical Methods in
  Natural Language Processing (EMNLP)}, pages 295--302, Online. Association for
  Computational Linguistics.

\bibitem[{Dhuliawala et~al.(2023)Dhuliawala, Zouhar, El-Assady, and
  Sachan}]{dhuliawala-etal-2023-diachronic}
Shehzaad Dhuliawala, Vil{\'e}m Zouhar, Mennatallah El-Assady, and Mrinmaya
  Sachan. 2023.
\newblock \href {https://doi.org/10.18653/v1/2023.emnlp-main.339} {A diachronic
  perspective on user trust in {AI} under uncertainty}.
\newblock In \emph{Proceedings of the 2023 Conference on Empirical Methods in
  Natural Language Processing}, pages 5567--5580, Singapore. Association for
  Computational Linguistics.

\bibitem[{Dong et~al.(2018)Dong, Quirk, and Lapata}]{dong-etal-2018-confidence}
Li~Dong, Chris Quirk, and Mirella Lapata. 2018.
\newblock \href {https://doi.org/10.18653/v1/P18-1069} {Confidence modeling for
  neural semantic parsing}.
\newblock In \emph{Proceedings of the 56th Annual Meeting of the Association
  for Computational Linguistics (Volume 1: Long Papers)}, pages 743--753,
  Melbourne, Australia. Association for Computational Linguistics.

\bibitem[{Dou et~al.(2022)Dou, Forbes, Koncel-Kedziorski, Smith, and
  Choi}]{dou-etal-2022-gpt}
Yao Dou, Maxwell Forbes, Rik Koncel-Kedziorski, Noah~A. Smith, and Yejin Choi.
  2022.
\newblock \href {https://doi.org/10.18653/v1/2022.acl-long.501} {Is {GPT}-3
  text indistinguishable from human text? scarecrow: A framework for
  scrutinizing machine text}.
\newblock In \emph{Proceedings of the 60th Annual Meeting of the Association
  for Computational Linguistics (Volume 1: Long Papers)}, pages 7250--7274,
  Dublin, Ireland. Association for Computational Linguistics.

\bibitem[{Fomicheva et~al.(2020)Fomicheva, Sun, Yankovskaya, Blain, Guzm{\'a}n,
  Fishel, Aletras, Chaudhary, and Specia}]{fomicheva-etal-2020-unsupervised}
Marina Fomicheva, Shuo Sun, Lisa Yankovskaya, Fr{\'e}d{\'e}ric Blain, Francisco
  Guzm{\'a}n, Mark Fishel, Nikolaos Aletras, Vishrav Chaudhary, and Lucia
  Specia. 2020.
\newblock \href {https://doi.org/10.1162/tacl_a_00330} {Unsupervised quality
  estimation for neural machine translation}.
\newblock \emph{Transactions of the Association for Computational Linguistics},
  8:539--555.

\bibitem[{Fornaciari et~al.(2021)Fornaciari, Uma, Paun, Plank, Hovy, and
  Poesio}]{fornaciari-etal-2021-beyond}
Tommaso Fornaciari, Alexandra Uma, Silviu Paun, Barbara Plank, Dirk Hovy, and
  Massimo Poesio. 2021.
\newblock \href {https://doi.org/10.18653/v1/2021.naacl-main.204} {Beyond black
  {\&} white: Leveraging annotator disagreement via soft-label multi-task
  learning}.
\newblock In \emph{Proceedings of the 2021 Conference of the North American
  Chapter of the Association for Computational Linguistics: Human Language
  Technologies}, pages 2591--2597, Online. Association for Computational
  Linguistics.

\bibitem[{Gan et~al.(2017)Gan, Li, Chen, Pu, Su, and
  Carin}]{gan-etal-2017-scalable}
Zhe Gan, Chunyuan Li, Changyou Chen, Yunchen Pu, Qinliang Su, and Lawrence
  Carin. 2017.
\newblock \href {https://doi.org/10.18653/v1/P17-1030} {Scalable {B}ayesian
  learning of recurrent neural networks for language modeling}.
\newblock In \emph{Proceedings of the 55th Annual Meeting of the Association
  for Computational Linguistics (Volume 1: Long Papers)}, pages 321--331,
  Vancouver, Canada. Association for Computational Linguistics.

\bibitem[{Geva et~al.(2019)Geva, Goldberg, and
  Berant}]{geva-etal-2019-modeling}
Mor Geva, Yoav Goldberg, and Jonathan Berant. 2019.
\newblock \href {https://doi.org/10.18653/v1/D19-1107} {Are we modeling the
  task or the annotator? an investigation of annotator bias in natural language
  understanding datasets}.
\newblock In \emph{Proceedings of the 2019 Conference on Empirical Methods in
  Natural Language Processing and the 9th International Joint Conference on
  Natural Language Processing (EMNLP-IJCNLP)}, pages 1161--1166, Hong Kong,
  China. Association for Computational Linguistics.

\bibitem[{Giovannotti and Gammerman(2021)}]{pmlr-v152-giovannotti21a}
Patrizio Giovannotti and Alex Gammerman. 2021.
\newblock \href {https://proceedings.mlr.press/v152/giovannotti21a.html}
  {Transformer-based conformal predictors for paraphrase detection}.
\newblock In \emph{Proceedings of the Tenth Symposium on Conformal and
  Probabilistic Prediction and Applications}, volume 152 of \emph{Proceedings
  of Machine Learning Research}, pages 243--265. PMLR.

\bibitem[{Glockner et~al.(2023)Glockner, Staliūnaitė, Thorne, Vallejo,
  Vlachos, and Gurevych}]{glockner2023ambifc}
Max Glockner, Ieva Staliūnaitė, James Thorne, Gisela Vallejo, Andreas
  Vlachos, and Iryna Gurevych. 2023.
\newblock \href {https://arxiv.org/abs/2104.00640} {Ambifc: Fact-checking
  ambiguous claims with evidence}.
\newblock \emph{arXiv preprint arXiv:2104.00640}.

\bibitem[{Glushkova et~al.(2021)Glushkova, Zerva, Rei, and
  Martins}]{glushkova-etal-2021-uncertainty-aware}
Taisiya Glushkova, Chrysoula Zerva, Ricardo Rei, and Andr{\'e} F.~T. Martins.
  2021.
\newblock \href {https://doi.org/10.18653/v1/2021.findings-emnlp.330}
  {Uncertainty-aware machine translation evaluation}.
\newblock In \emph{Findings of the Association for Computational Linguistics:
  EMNLP 2021}, pages 3920--3938, Punta Cana, Dominican Republic. Association
  for Computational Linguistics.

\bibitem[{Goldberg and Hirst(2017)}]{neural-goldberg-hirst-2017}
Yoav Goldberg and Graeme Hirst. 2017.
\newblock \emph{Neural Network Methods in Natural Language Processing}.
\newblock Morgan \& Claypool Publishers.

\bibitem[{Gordon et~al.(2022)Gordon, Lam, Park, Patel, Hancock, Hashimoto, and
  Bernstein}]{gordon2022jury}
Mitchell~L Gordon, Michelle~S Lam, Joon~Sung Park, Kayur Patel, Jeff Hancock,
  Tatsunori Hashimoto, and Michael~S Bernstein. 2022.
\newblock Jury learning: Integrating dissenting voices into machine learning
  models.
\newblock In \emph{Proceedings of the 2022 CHI Conference on Human Factors in
  Computing Systems}, pages 1--19.

\bibitem[{Gordon et~al.(2021)Gordon, Zhou, Patel, Hashimoto, and
  Bernstein}]{gordon-etal-2021-disagreement}
Mitchell~L. Gordon, Kaitlyn Zhou, Kayur Patel, Tatsunori Hashimoto, and
  Michael~S. Bernstein. 2021.
\newblock \href {https://doi.org/10.1145/3411764.3445423} {The disagreement
  deconvolution: Bringing machine learning performance metrics in line with
  reality}.
\newblock In \emph{Association for Computing Machinery}, CHI '21, New York, NY,
  USA.

\bibitem[{Gruber et~al.(2023)Gruber, Schenk, Schierholz, Kreuter, and
  Kauermann}]{gruber2023sources}
Cornelia Gruber, Patrick~Oliver Schenk, Malte Schierholz, Frauke Kreuter, and
  G{\"o}ran Kauermann. 2023.
\newblock Sources of uncertainty in machine learning--a statisticians' view.
\newblock \emph{arXiv preprint arXiv:2305.16703}.

\bibitem[{Guo et~al.(2017)Guo, Pleiss, Sun, and
  Weinberger}]{guo2017calibration}
Chuan Guo, Geoff Pleiss, Yu~Sun, and Kilian~Q Weinberger. 2017.
\newblock On calibration of modern neural networks.
\newblock In \emph{International conference on machine learning}, pages
  1321--1330. PMLR.

\bibitem[{Gupta et~al.(2021)Gupta, Rahimi, Ajanthan, Mensink, Sminchisescu, and
  Hartley}]{gupta2021calibration}
Kartik Gupta, Amir Rahimi, Thalaiyasingam Ajanthan, Thomas Mensink, Cristian
  Sminchisescu, and Richard Hartley. 2021.
\newblock \href {https://openreview.net/forum?id=eQe8DEWNN2W} {Calibration of
  neural networks using splines}.
\newblock In \emph{International Conference on Learning Representations}.

\bibitem[{Halpern(2017)}]{halpern2017reasoning}
Joseph~Y Halpern. 2017.
\newblock \emph{Reasoning about uncertainty}.
\newblock MIT press.

\bibitem[{H{\"u}llermeier and Waegeman(2021)}]{hullermeier2021aleatoric}
Eyke H{\"u}llermeier and Willem Waegeman. 2021.
\newblock Aleatoric and epistemic uncertainty in machine learning: An
  introduction to concepts and methods.
\newblock \emph{Machine Learning}, 110:457--506.

\bibitem[{Jiang and Marneffe(2022)}]{jiang2022investigating}
Nan-Jiang Jiang and Marie-Catherine~de Marneffe. 2022.
\newblock Investigating reasons for disagreement in natural language inference.
\newblock \emph{Transactions of the Association for Computational Linguistics},
  10:1357--1374.

\bibitem[{Jiang et~al.(2023)Jiang, Tan, and
  de~Marneffe}]{jiang2023understanding}
Nan-Jiang Jiang, Chenhao Tan, and Marie-Catherine de~Marneffe. 2023.
\newblock Understanding and predicting human label variation in natural
  language inference through explanation.
\newblock \emph{arXiv preprint arXiv:2304.12443}.

\bibitem[{Jiang et~al.(2021{\natexlab{a}})Jiang, Araki, Ding, and
  Neubig}]{jiang2021can}
Zhengbao Jiang, Jun Araki, Haibo Ding, and Graham Neubig. 2021{\natexlab{a}}.
\newblock How can we know when language models know? on the calibration of
  language models for question answering.
\newblock \emph{Transactions of the Association for Computational Linguistics},
  9:962--977.

\bibitem[{Jiang et~al.(2021{\natexlab{b}})Jiang, Araki, Ding, and
  Neubig}]{jiang-etal-2021-know}
Zhengbao Jiang, Jun Araki, Haibo Ding, and Graham Neubig. 2021{\natexlab{b}}.
\newblock \href {https://doi.org/10.1162/tacl_a_00407} {How can we know when
  language models know? on the calibration of language models for question
  answering}.
\newblock \emph{Transactions of the Association for Computational Linguistics},
  9:962--977.

\bibitem[{Kamath et~al.(2020)Kamath, Jia, and
  Liang}]{kamath-etal-2020-selective}
Amita Kamath, Robin Jia, and Percy Liang. 2020.
\newblock \href {https://doi.org/10.18653/v1/2020.acl-main.503} {Selective
  question answering under domain shift}.
\newblock In \emph{Proceedings of the 58th Annual Meeting of the Association
  for Computational Linguistics}, pages 5684--5696, Online. Association for
  Computational Linguistics.

\bibitem[{Kennedy et~al.(2022)Kennedy, Atari, Davani, Yeh, Omrani, Kim, Coombs,
  Havaldar, Portillo-Wightman, Gonzalez et~al.}]{kennedy2022introducing}
Brendan Kennedy, Mohammad Atari, Aida~Mostafazadeh Davani, Leigh Yeh, Ali
  Omrani, Yehsong Kim, Kris Coombs, Shreya Havaldar, Gwenyth Portillo-Wightman,
  Elaine Gonzalez, et~al. 2022.
\newblock Introducing the gab hate corpus: defining and applying hate-based
  rhetoric to social media posts at scale.
\newblock \emph{Language Resources and Evaluation}, pages 1--30.

\bibitem[{Klie et~al.(2022)Klie, Webber, and Gurevych}]{klie2022annotation}
Jan-Christoph Klie, Bonnie Webber, and Iryna Gurevych. 2022.
\newblock Annotation error detection: Analyzing the past and present for a more
  coherent future.
\newblock \emph{Computational Linguistics}, pages 1--42.

\bibitem[{Kolmogorov(1960)}]{kolmogorov1960foundations}
Andrey~N. Kolmogorov. 1960.
\newblock \emph{Foundations of the Theory of Probability}, 2 edition.
\newblock Chelsea Pub Co.

\bibitem[{Kong et~al.(2020)Kong, Jiang, Zhuang, Lyu, Zhao, and
  Zhang}]{kong-etal-2020-calibrated}
Lingkai Kong, Haoming Jiang, Yuchen Zhuang, Jie Lyu, Tuo Zhao, and Chao Zhang.
  2020.
\newblock \href {https://doi.org/10.18653/v1/2020.emnlp-main.102} {Calibrated
  language model fine-tuning for in- and out-of-distribution data}.
\newblock In \emph{Proceedings of the 2020 Conference on Empirical Methods in
  Natural Language Processing (EMNLP)}, pages 1326--1340, Online. Association
  for Computational Linguistics.

\bibitem[{Kull et~al.(2019)Kull, Perello~Nieto, K{\"a}ngsepp, Silva~Filho,
  Song, and Flach}]{kull2019beyond}
Meelis Kull, Miquel Perello~Nieto, Markus K{\"a}ngsepp, Telmo Silva~Filho, Hao
  Song, and Peter Flach. 2019.
\newblock Beyond temperature scaling: Obtaining well-calibrated multi-class
  probabilities with dirichlet calibration.
\newblock \emph{Advances in neural information processing systems}, 32.

\bibitem[{Kumar et~al.(2019)Kumar, Liang, and Ma}]{kumar2019verified}
Ananya Kumar, Percy~S Liang, and Tengyu Ma. 2019.
\newblock Verified uncertainty calibration.
\newblock \emph{Advances in Neural Information Processing Systems}, 32.

\bibitem[{Li et~al.(2022)Li, Hu, and Chen}]{li-etal-2022-calibration}
Dongfang Li, Baotian Hu, and Qingcai Chen. 2022.
\newblock \href {https://doi.org/10.18653/v1/2022.emnlp-main.178} {Calibration
  meets explanation: A simple and effective approach for model confidence
  estimates}.
\newblock In \emph{Proceedings of the 2022 Conference on Empirical Methods in
  Natural Language Processing}, pages 2775--2784, Abu Dhabi, United Arab
  Emirates. Association for Computational Linguistics.

\bibitem[{Lindley(2013)}]{lindley2013understanding}
Dennis~V Lindley. 2013.
\newblock \emph{Understanding uncertainty}.
\newblock John Wiley \& Sons.

\bibitem[{Liu et~al.(2023)Liu, Wu, Michael, Suhr, West, Koller, Swayamdipta,
  Smith, and Choi}]{liu2023we}
Alisa Liu, Zhaofeng Wu, Julian Michael, Alane Suhr, Peter West, Alexander
  Koller, Swabha Swayamdipta, Noah~A Smith, and Yejin Choi. 2023.
\newblock We're afraid language models aren't modeling ambiguity.
\newblock \emph{arXiv preprint arXiv:2304.14399}.

\bibitem[{Lovchinsky et~al.(2020)Lovchinsky, Daks, Malkin, Samangouei, Saeedi,
  Liu, Sankaranarayanan, Gafner, Sternlieb, Maher
  et~al.}]{lovchinsky2020discrepancy}
Igor Lovchinsky, Alon Daks, Israel Malkin, Pouya Samangouei, Ardavan Saeedi,
  Yang Liu, Swami Sankaranarayanan, Tomer Gafner, Ben Sternlieb, Patrick Maher,
  et~al. 2020.
\newblock Discrepancy ratio: Evaluating model performance when even experts
  disagree on the truth.
\newblock In \emph{International Conference on Learning Representations}.

\bibitem[{Maltoudoglou et~al.(2020)Maltoudoglou, Paisios, and
  Papadopoulos}]{pmlr-v128-maltoudoglou20a}
Lysimachos Maltoudoglou, Andreas Paisios, and Harris Papadopoulos. 2020.
\newblock \href {https://proceedings.mlr.press/v128/maltoudoglou20a.html}
  {Bert-based conformal predictor for sentiment analysis}.
\newblock In \emph{Proceedings of the Ninth Symposium on Conformal and
  Probabilistic Prediction and Applications}, volume 128 of \emph{Proceedings
  of Machine Learning Research}, pages 269--284. PMLR.

\bibitem[{Manning(2011)}]{manning2011part}
Christopher~D Manning. 2011.
\newblock Part-of-speech tagging from 97\% to 100\%: is it time for some
  linguistics?
\newblock In \emph{Computational Linguistics and Intelligent Text Processing:
  12th International Conference, CICLing 2011, Tokyo, Japan, February 20-26,
  2011. Proceedings, Part I 12}, pages 171--189. Springer.

\bibitem[{Meissner et~al.(2021)Meissner, Thumwanit, Sugawara, and
  Aizawa}]{Meissner-etal-2021-embracing}
Johannes~Mario Meissner, Napat Thumwanit, Saku Sugawara, and Akiko Aizawa.
  2021.
\newblock \href {https://doi.org/10.18653/v1/2021.acl-short.109} {Embracing
  ambiguity: {S}hifting the training target of {NLI} models}.
\newblock In \emph{Proceedings of the 59th Annual Meeting of the Association
  for Computational Linguistics and the 11th International Joint Conference on
  Natural Language Processing (Volume 2: Short Papers)}, pages 862--869,
  Online. Association for Computational Linguistics.

\bibitem[{Min et~al.(2020)Min, Michael, Hajishirzi, and
  Zettlemoyer}]{min-etal-2020-ambigqa}
Sewon Min, Julian Michael, Hannaneh Hajishirzi, and Luke Zettlemoyer. 2020.
\newblock \href {https://doi.org/10.18653/v1/2020.emnlp-main.466} {{A}mbig{QA}:
  Answering ambiguous open-domain questions}.
\newblock In \emph{Proceedings of the 2020 Conference on Empirical Methods in
  Natural Language Processing (EMNLP)}, pages 5783--5797, Online. Association
  for Computational Linguistics.

\bibitem[{Mostafazadeh~Davani et~al.(2022)Mostafazadeh~Davani, D{\'\i}az, and
  Prabhakaran}]{davani-etal-2022-dealing}
Aida Mostafazadeh~Davani, Mark D{\'\i}az, and Vinodkumar Prabhakaran. 2022.
\newblock \href {https://doi.org/10.1162/tacl_a_00449} {Dealing with
  disagreements: Looking beyond the majority vote in subjective annotations}.
\newblock \emph{Transactions of the Association for Computational Linguistics},
  10:92--110.

\bibitem[{Mukhoti et~al.(2023)Mukhoti, Kirsch, van Amersfoort, Torr, and
  Gal}]{mukhoti2023deep}
Jishnu Mukhoti, Andreas Kirsch, Joost van Amersfoort, Philip~HS Torr, and Yarin
  Gal. 2023.
\newblock Deep deterministic uncertainty: A new simple baseline.
\newblock In \emph{Proceedings of the IEEE/CVF Conference on Computer Vision
  and Pattern Recognition}, pages 24384--24394.

\bibitem[{Naeini et~al.(2015)Naeini, Cooper, and
  Hauskrecht}]{naeini2015obtaining}
Mahdi~Pakdaman Naeini, Gregory Cooper, and Milos Hauskrecht. 2015.
\newblock Obtaining well calibrated probabilities using bayesian binning.
\newblock In \emph{Proceedings of the AAAI conference on artificial
  intelligence}, volume~29.

\bibitem[{Nie et~al.(2020)Nie, Zhou, and Bansal}]{nie-etal-2020-learn}
Yixin Nie, Xiang Zhou, and Mohit Bansal. 2020.
\newblock \href {https://doi.org/10.18653/v1/2020.emnlp-main.734} {What can we
  learn from collective human opinions on natural language inference data?}
\newblock In \emph{Proceedings of the 2020 Conference on Empirical Methods in
  Natural Language Processing (EMNLP)}, pages 9131--9143, Online. Association
  for Computational Linguistics.

\bibitem[{Nixon et~al.(2019)Nixon, Dusenberry, Zhang, Jerfel, and
  Tran}]{nixon2019measuring}
Jeremy Nixon, Michael~W Dusenberry, Linchuan Zhang, Ghassen Jerfel, and Dustin
  Tran. 2019.
\newblock Measuring calibration in deep learning.
\newblock In \emph{CVPR workshops}, 7.

\bibitem[{Park and Caragea(2022)}]{park-caragea-2022-calibration}
Seo~Yeon Park and Cornelia Caragea. 2022.
\newblock \href {https://doi.org/10.18653/v1/2022.acl-long.368} {On the
  calibration of pre-trained language models using mixup guided by area under
  the margin and saliency}.
\newblock In \emph{Proceedings of the 60th Annual Meeting of the Association
  for Computational Linguistics (Volume 1: Long Papers)}, pages 5364--5374,
  Dublin, Ireland. Association for Computational Linguistics.

\bibitem[{Paun et~al.(2018)Paun, Carpenter, Chamberlain, Hovy, Kruschwitz, and
  Poesio}]{paun-etal-2018-comparing}
Silviu Paun, Bob Carpenter, Jon Chamberlain, Dirk Hovy, Udo Kruschwitz, and
  Massimo Poesio. 2018.
\newblock \href {https://doi.org/10.1162/tacl_a_00040} {Comparing {B}ayesian
  models of annotation}.
\newblock \emph{Transactions of the Association for Computational Linguistics},
  6:571--585.

\bibitem[{Pavlick and Kwiatkowski(2019)}]{pavlick-kwiatkowski-2019-inherent}
Ellie Pavlick and Tom Kwiatkowski. 2019.
\newblock \href {https://doi.org/10.1162/tacl_a_00293} {Inherent disagreements
  in human textual inferences}.
\newblock \emph{Transactions of the Association for Computational Linguistics},
  7:677--694.

\bibitem[{Pei et~al.(2022)Pei, Wang, and Szarvas}]{pei2022transformer}
Jiahuan Pei, Cheng Wang, and Gy{\"o}rgy Szarvas. 2022.
\newblock Transformer uncertainty estimation with hierarchical stochastic
  attention.
\newblock In \emph{Proceedings of the AAAI Conference on Artificial
  Intelligence}, volume~36, pages 11147--11155.

\bibitem[{Peterson et~al.(2019)Peterson, Battleday, Griffiths, and
  Russakovsky}]{peterson2019human}
Joshua~C Peterson, Ruairidh~M Battleday, Thomas~L Griffiths, and Olga
  Russakovsky. 2019.
\newblock Human uncertainty makes classification more robust.
\newblock In \emph{Proceedings of the IEEE/CVF International Conference on
  Computer Vision}, pages 9617--9626.

\bibitem[{Plank(2022)}]{plank-2022-problem}
Barbara Plank. 2022.
\newblock \href {https://doi.org/10.18653/v1/2022.emnlp-main.731} {The
  {``}problem{''} of human label variation: On ground truth in data, modeling
  and evaluation}.
\newblock In \emph{Proceedings of the 2022 Conference on Empirical Methods in
  Natural Language Processing}, pages 10671--10682, Abu Dhabi, United Arab
  Emirates. Association for Computational Linguistics.

\bibitem[{Plank et~al.(2014)Plank, Hovy, and
  S{\o}gaard}]{plank-etal-2014-linguistically}
Barbara Plank, Dirk Hovy, and Anders S{\o}gaard. 2014.
\newblock \href {https://doi.org/10.3115/v1/P14-2083} {Linguistically debatable
  or just plain wrong?}
\newblock In \emph{Proceedings of the 52nd Annual Meeting of the Association
  for Computational Linguistics (Volume 2: Short Papers)}, pages 507--511,
  Baltimore, Maryland. Association for Computational Linguistics.

\bibitem[{Poesio and Artstein(2005)}]{poesio-artstein-2005-reliability}
Massimo Poesio and Ron Artstein. 2005.
\newblock \href {https://aclanthology.org/W05-0311} {The reliability of
  anaphoric annotation, reconsidered: Taking ambiguity into account}.
\newblock In \emph{Proceedings of the Workshop on Frontiers in Corpus
  Annotations {II}: Pie in the Sky}, pages 76--83, Ann Arbor, Michigan.
  Association for Computational Linguistics.

\bibitem[{Ponti et~al.(2021)Ponti, Vuli{\'c}, Cotterell, Parovic, Reichart, and
  Korhonen}]{ponti-etal-2021-parameter}
Edoardo~M. Ponti, Ivan Vuli{\'c}, Ryan Cotterell, Marinela Parovic, Roi
  Reichart, and Anna Korhonen. 2021.
\newblock \href {https://doi.org/10.1162/tacl_a_00374} {Parameter space
  factorization for zero-shot learning across tasks and languages}.
\newblock \emph{Transactions of the Association for Computational Linguistics},
  9:410--428.

\bibitem[{Raykar and Yu(2011)}]{raykar2011ranking}
Vikas~C Raykar and Shipeng Yu. 2011.
\newblock Ranking annotators for crowdsourced labeling tasks.
\newblock \emph{Advances in neural information processing systems}, 24.

\bibitem[{Rottger et~al.(2022)Rottger, Vidgen, Hovy, and
  Pierrehumbert}]{rottger-etal-2022-two}
Paul Rottger, Bertie Vidgen, Dirk Hovy, and Janet Pierrehumbert. 2022.
\newblock \href {https://doi.org/10.18653/v1/2022.naacl-main.13} {Two
  contrasting data annotation paradigms for subjective {NLP} tasks}.
\newblock In \emph{Proceedings of the 2022 Conference of the North American
  Chapter of the Association for Computational Linguistics: Human Language
  Technologies}, pages 175--190, Seattle, United States. Association for
  Computational Linguistics.

\bibitem[{Scholman et~al.(2022)Scholman, Dong, Yung, and
  Demberg}]{scholman-etal-2022-discogem}
Merel Scholman, Tianai Dong, Frances Yung, and Vera Demberg. 2022.
\newblock \href {https://aclanthology.org/2022.lrec-1.351} {{D}isco{G}e{M}: A
  crowdsourced corpus of genre-mixed implicit discourse relations}.
\newblock In \emph{Proceedings of the Thirteenth Language Resources and
  Evaluation Conference}, pages 3281--3290, Marseille, France. European
  Language Resources Association.

\bibitem[{Shareghi et~al.(2019)Shareghi, Li, Zhu, Reichart, and
  Korhonen}]{shareghi-etal-2019-bayesian}
Ehsan Shareghi, Yingzhen Li, Yi~Zhu, Roi Reichart, and Anna Korhonen. 2019.
\newblock \href {https://doi.org/10.18653/v1/N19-1354} {{B}ayesian learning for
  neural dependency parsing}.
\newblock In \emph{Proceedings of the 2019 Conference of the North {A}merican
  Chapter of the Association for Computational Linguistics: Human Language
  Technologies, Volume 1 (Long and Short Papers)}, pages 3509--3519,
  Minneapolis, Minnesota. Association for Computational Linguistics.

\bibitem[{Shelmanov et~al.(2021)Shelmanov, Tsymbalov, Puzyrev, Fedyanin,
  Panchenko, and Panov}]{shelmanov-etal-2021-certain}
Artem Shelmanov, Evgenii Tsymbalov, Dmitri Puzyrev, Kirill Fedyanin, Alexander
  Panchenko, and Maxim Panov. 2021.
\newblock \href {https://doi.org/10.18653/v1/2021.eacl-main.157} {How certain
  is your {T}ransformer?}
\newblock In \emph{Proceedings of the 16th Conference of the European Chapter
  of the Association for Computational Linguistics: Main Volume}, pages
  1833--1840, Online. Association for Computational Linguistics.

\bibitem[{Silberer et~al.(2020)Silberer, Zarrie{\ss}, Westera, and
  Boleda}]{silberer-etal-2020-humans}
Carina Silberer, Sina Zarrie{\ss}, Matthijs Westera, and Gemma Boleda. 2020.
\newblock \href {https://doi.org/10.18653/v1/2020.coling-main.172} {Humans meet
  models on object naming: A new dataset and analysis}.
\newblock In \emph{Proceedings of the 28th International Conference on
  Computational Linguistics}, pages 1893--1905, Barcelona, Spain (Online).
  International Committee on Computational Linguistics.

\bibitem[{Specia et~al.(2009)Specia, Turchi, Cancedda, Cristianini, and
  Dymetman}]{specia-etal-2009-estimating}
Lucia Specia, Marco Turchi, Nicola Cancedda, Nello Cristianini, and Marc
  Dymetman. 2009.
\newblock \href {https://aclanthology.org/2009.eamt-1.5} {Estimating the
  sentence-level quality of machine translation systems}.
\newblock In \emph{Proceedings of the 13th Annual Conference of the European
  Association for Machine Translation}, Barcelona, Spain. European Association
  for Machine Translation.

\bibitem[{Ulmer et~al.(2022)Ulmer, Frellsen, and
  Hardmeier}]{ulmer-etal-2022-exploring}
Dennis Ulmer, Jes Frellsen, and Christian Hardmeier. 2022.
\newblock \href {https://doi.org/10.18653/v1/2022.findings-emnlp.198}
  {Exploring predictive uncertainty and calibration in {NLP}: A study on the
  impact of method {\&} data scarcity}.
\newblock In \emph{Findings of the Association for Computational Linguistics:
  EMNLP 2022}, pages 2707--2735, Abu Dhabi, United Arab Emirates. Association
  for Computational Linguistics.

\bibitem[{Uma et~al.(2020)Uma, Fornaciari, Hovy, Paun, Plank, and
  Poesio}]{uma2020case}
Alexandra Uma, Tommaso Fornaciari, Dirk Hovy, Silviu Paun, Barbara Plank, and
  Massimo Poesio. 2020.
\newblock A case for soft loss functions.
\newblock In \emph{Proceedings of the AAAI Conference on Human Computation and
  Crowdsourcing}, volume~8, pages 173--177.

\bibitem[{Uma et~al.(2021)Uma, Fornaciari, Hovy, Paun, Plank, and
  Poesio}]{uma2021learning}
Alexandra~N Uma, Tommaso Fornaciari, Dirk Hovy, Silviu Paun, Barbara Plank, and
  Massimo Poesio. 2021.
\newblock Learning from disagreement: A survey.
\newblock \emph{Journal of Artificial Intelligence Research}, 72:1385--1470.

\bibitem[{Vaicenavicius et~al.(2019)Vaicenavicius, Widmann, Andersson,
  Lindsten, Roll, and Sch{\"o}n}]{vaicenavicius2019evaluating}
Juozas Vaicenavicius, David Widmann, Carl Andersson, Fredrik Lindsten, Jacob
  Roll, and Thomas Sch{\"o}n. 2019.
\newblock Evaluating model calibration in classification.
\newblock In \emph{The 22nd International Conference on Artificial Intelligence
  and Statistics}, pages 3459--3467. PMLR.

\bibitem[{Van~Amersfoort et~al.(2020)Van~Amersfoort, Smith, Teh, and
  Gal}]{van2020uncertainty}
Joost Van~Amersfoort, Lewis Smith, Yee~Whye Teh, and Yarin Gal. 2020.
\newblock Uncertainty estimation using a single deep deterministic neural
  network.
\newblock In \emph{International conference on machine learning}, pages
  9690--9700. PMLR.

\bibitem[{Vasconcelos et~al.(2023)Vasconcelos, Bansal, Fourney, Liao, and
  Vaughan}]{vasconcelos2023generation}
Helena Vasconcelos, Gagan Bansal, Adam Fourney, Q~Vera Liao, and
  Jennifer~Wortman Vaughan. 2023.
\newblock Generation probabilities are not enough: Exploring the effectiveness
  of uncertainty highlighting in ai-powered code completions.
\newblock \emph{arXiv preprint arXiv:2302.07248}.

\bibitem[{Vazhentsev et~al.(2022)Vazhentsev, Kuzmin, Shelmanov, Tsvigun,
  Tsymbalov, Fedyanin, Panov, Panchenko, Gusev, Burtsev, Avetisian, and
  Zhukov}]{vazhentsev-etal-2022-uncertainty}
Artem Vazhentsev, Gleb Kuzmin, Artem Shelmanov, Akim Tsvigun, Evgenii
  Tsymbalov, Kirill Fedyanin, Maxim Panov, Alexander Panchenko, Gleb Gusev,
  Mikhail Burtsev, Manvel Avetisian, and Leonid Zhukov. 2022.
\newblock \href {https://doi.org/10.18653/v1/2022.acl-long.566} {Uncertainty
  estimation of transformer predictions for misclassification detection}.
\newblock In \emph{Proceedings of the 60th Annual Meeting of the Association
  for Computational Linguistics (Volume 1: Long Papers)}, pages 8237--8252,
  Dublin, Ireland. Association for Computational Linguistics.

\bibitem[{Vodrahalli et~al.(2022)Vodrahalli, Gerstenberg, and
  Zou}]{vodrahalli2022uncalibrated}
Kailas Vodrahalli, Tobias Gerstenberg, and James~Y Zou. 2022.
\newblock Uncalibrated models can improve human-ai collaboration.
\newblock \emph{Advances in Neural Information Processing Systems},
  35:4004--4016.

\bibitem[{Wang and Yin(2021)}]{wang-yin-2021-explanations}
Xinru Wang and Ming Yin. 2021.
\newblock \href {https://doi.org/10.1145/3397481.3450650} {Are explanations
  helpful? a comparative study of the effects of explanations in ai-assisted
  decision-making}.
\newblock In \emph{26th International Conference on Intelligent User
  Interfaces}, IUI '21, page 318–328, New York, NY, USA. Association for
  Computing Machinery.

\bibitem[{Weber and Plank(2023)}]{weber-plank-2023-activeaed}
Leon Weber and Barbara Plank. 2023.
\newblock \href {https://doi.org/10.18653/v1/2023.findings-acl.562}
  {{A}ctive{AED}: A human in the loop improves annotation error detection}.
\newblock In \emph{Findings of the Association for Computational Linguistics:
  ACL 2023}, pages 8834--8845, Toronto, Canada. Association for Computational
  Linguistics.

\bibitem[{Wei et~al.(2022)Wei, Zhu, Cheng, Liu, Niu, and Liu}]{wei2022learning}
Jiaheng Wei, Zhaowei Zhu, Hao Cheng, Tongliang Liu, Gang Niu, and Yang Liu.
  2022.
\newblock \href {https://openreview.net/forum?id=TBWA6PLJZQm} {Learning with
  noisy labels revisited: A study using real-world human annotations}.
\newblock In \emph{International Conference on Learning Representations}.

\bibitem[{Widmann et~al.(2019)Widmann, Lindsten, and
  Zachariah}]{widmann2019calibration}
David Widmann, Fredrik Lindsten, and Dave Zachariah. 2019.
\newblock Calibration tests in multi-class classification: A unifying
  framework.
\newblock \emph{Advances in neural information processing systems}, 32.

\bibitem[{Ye and Durrett(2022)}]{ye-durrett-2022-explanations}
Xi~Ye and Greg Durrett. 2022.
\newblock \href {https://doi.org/10.18653/v1/2022.acl-long.429} {Can
  explanations be useful for calibrating black box models?}
\newblock In \emph{Proceedings of the 60th Annual Meeting of the Association
  for Computational Linguistics (Volume 1: Long Papers)}, pages 6199--6212,
  Dublin, Ireland. Association for Computational Linguistics.

\bibitem[{Yoshikawa and Okazaki(2023)}]{yoshikawa2023selective}
Hiyori Yoshikawa and Naoaki Okazaki. 2023.
\newblock Selective-lama: Selective prediction for confidence-aware evaluation
  of language models.
\newblock In \emph{Findings of the Association for Computational Linguistics:
  EACL 2023}, pages 1972--1983.

\bibitem[{Zerva and Martins(2023)}]{zerva2023conformalizing}
Chrysoula Zerva and Andr{\'e}~FT Martins. 2023.
\newblock Conformalizing machine translation evaluation.
\newblock \emph{arXiv preprint arXiv:2306.06221}.

\bibitem[{Zhang et~al.(2021)Zhang, Gong, and
  Choi}]{zhang-etal-2021-learning-different}
Shujian Zhang, Chengyue Gong, and Eunsol Choi. 2021.
\newblock \href {https://doi.org/10.18653/v1/2021.emnlp-main.601} {Learning
  with different amounts of annotation: From zero to many labels}.
\newblock In \emph{Proceedings of the 2021 Conference on Empirical Methods in
  Natural Language Processing}, pages 7620--7632, Online and Punta Cana,
  Dominican Republic. Association for Computational Linguistics.

\bibitem[{Zhang et~al.(2020)Zhang, Liao, and Bellamy}]{zhang-etal-2020-effect}
Yunfeng Zhang, Q.~Vera Liao, and Rachel K.~E. Bellamy. 2020.
\newblock \href {https://doi.org/10.1145/3351095.3372852} {Effect of confidence
  and explanation on accuracy and trust calibration in ai-assisted decision
  making}.
\newblock In \emph{Proceedings of the 2020 Conference on Fairness,
  Accountability, and Transparency}, FAT* '20, page 295–305, New York, NY,
  USA. Association for Computing Machinery.

\end{thebibliography}
\bibliographystyle{acl_natbib}

\appendix
\end{document}